# Investigating the Relationship between the Weighted Figure of Merit and Rosin's Measure

**BIMAL KUMAR RAY**
School of Computer Science Engineering and Information Systems
Vellore Institute of Technology, Vellore – 632014, INDIA

**ABSTRACT** Many studies have been conducted to solve the problem of approximating a digital boundary by piece straight-line segments for the further processing required in computer vision applications. The authors of these studies compared their schemes to determine the best one. The initial measure used to assess the goodness of fit of a polygonal approximation was the figure of merit. Later, it was noted that this measure was not an appropriate metric for a valid reason which is why Rosin – through mathematical analysis – introduced a measure called merit. However, this measure involves an optimal scheme of polygonal approximation, so it is time-consuming to compute it to assess the goodness of fit of an approximation. This led many researchers to use a weighted figure of merit as a substitute for Rosin's measure to compare sub optimal schemes. An attempt is made in this communication to investigate whether the two measures—weighted figure of merit and Rosin's measure—are related so that one can be used instead of the other, and toward this end, theoretical analysis, experimental investigation and statistical analysis are carried out. The mathematical formulas for the weighted figure of merit and Rosin's measure are analyzed, and through proof of theorems, it is found that the two measures are theoretically independent of each other. The graphical analysis of experiments carried out using a public dataset supports the results of the theoretical analysis. The statistical analysis via Pearson's correlation coefficient and non-linear correlation measure also revealed that the two measures are uncorrelated. This analysis leads one to conclude that if a suboptimal scheme is found to be better (worse) than some other suboptimal scheme, as indicated by Rosin's measure, then the same conclusion cannot be drawn using a weighted figure of merit, so one cannot use a weighted figure of merit instead of Rosin's measure.

**INDEX TERMS** Rosin's measure, weighted figure of merit, analysis, theoretical, experimental, statistical

## I INTRODUCTION

The boundary of a two-dimensional digital image can be represented by a sequence of digital coordinates determined by Freeman's eight-direction chain code. Typically, a large curve has too many points on its boundary; thus, the representation of a curved boundary by these points results in high storage and processing times for further analysis of a curve. It is better if a digital boundary is represented in a compact form, and one such means is to represent a boundary with fewer points than the total number of points the digital boundary has; this results in reduced storage and processing requirements. Polygonal approximation is one way of representing a curve with a reduced number of points. When a digital closed boundary is represented by a sequence of points that define the vertices of a polygon, the approximation is called a polygonal approximation. When an open digital curve is represented by a sequence of piecewise straight linear segments, the representation is called a polyline approximation. In this work, closed digital curves are considered, and the approximation considered is a polygonal approximation.

Several algorithms have been developed by researchers for approximating a digital boundary via a sequence of straight-line segments. Approximation algorithms in this area can be divided into two major categories: optimal and suboptimal. The optimal algorithms developed thus far include dynamic programming, the $A^*$ search algorithm, and mixed integer programming ([1], [2], [3], [4], [5]); however, these algorithms are computationally expensive. Suboptimal algorithms are more efficient than optimal algorithms; however, these algorithms are heuristic in nature. In addition to being classified as optimal or suboptimal, polygonal approximation techniques can be categorized as supervised or unsupervised approximations.

Supervised approximation requires human intervention to specify either the number of vertices required to represent the approximation or the error tolerance. Unsupervised approximation does not require human intervention; rather, it adaptively determines either the number of vertices or the approximation error on the basis of the implicit nature of a curve and the nature of the algorithm. Usually, the vertices of a polygonal approximation are a subset of the digital boundary points; however, there exists an approximation where the vertices are not forced to be a subset of the digital points, resulting in a more relaxed approximation, albeit at an additional cost.

It is necessary to use a quantitative measure to assess the quality of a polygonal approximation scheme. Initially, the *figure of merit,* defined by the ratio of the compression ratio to the sum of the squares of the errors, was introduced to measure the goodness of fit of an approximation. However, it was later found that this measure is inappropriate because of the imbalance between the two terms involved in the measure. An analytically derived measure of goodness is Rosin's measure, which uses an optimal scheme of polygonal approximation as the benchmark. However, it is time-consuming to assess an approximation via Rosin's measure because of the involvement of the optimal scheme. This is why many developers of polygonal approximation have used a variant of the *figure of merit* called the *weighted figure of merit* to assess suboptimal schemes of polygonal approximation. However, this article shows through analytical treatment supported by empirical results and statistical analysis that a weighted figure of merit cannot be a substitute for Rosin's measure because the two measures are independent—the behavior of one cannot determine the behavior of the other.

## II MEASURE OF THE GOODNESS OF POLYGON APPROXIMATION

A polygonal approximation of a digital curve is assessed via various measures, such as the compression ratio, maximum error and sum of the squares of the errors. A closed digital $(C)$ curve with $n$ points is defined by a circular sequence of $n$ digital points

$$C = \{p_i = (x_i, y_i): i = 1, .., n;\ p_{i\pm n} = p_i\}. \quad (1)$$

Any such curve can be approximated by a polygon with an arbitrary degree of accuracy via a supervised scheme of polygonal approximation, whereas an unsupervised scheme generates an approximation with accuracy determined by the implicit nature of a curve and the inherent characteristics of the approximation scheme. The figure below (Figure 1) shows a digital curve (left) and its polygonal approximation (right) via an unsupervised scheme. The vertices on the polygon are indicated with solid circles.

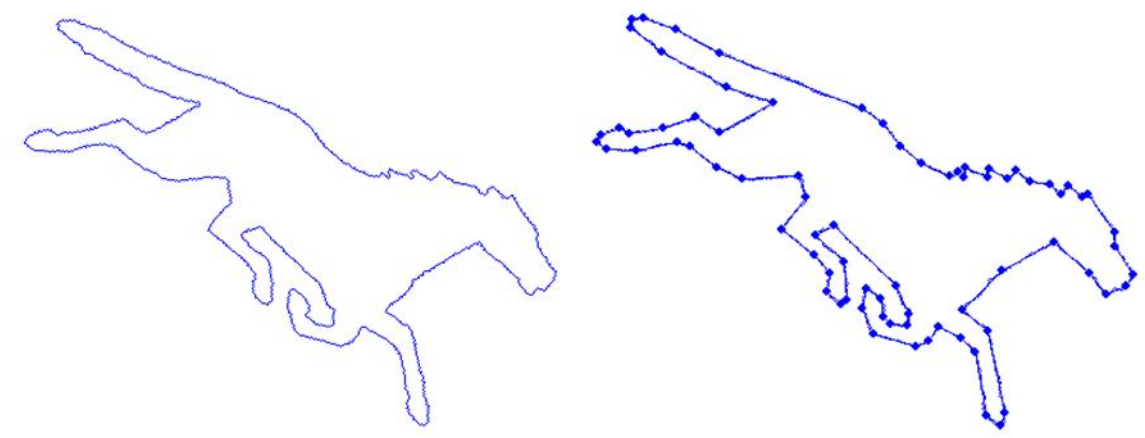

**FIGURE 1 A digital curve (left image) and its polygonal approximation (right image) via an unsupervised scheme. The vertices of the polygon are indicated with solid circles. The number of vertices on the right image is significantly less than the number of points represented as pixels in the left image.**

If a digital curve with $n$ points is approximated by a polygon with $m$ vertices, then the compression ratio$(CR)$ of the approximation is defined as

$$CR = \frac{n}{m} \quad (2)$$

The digital curve shown in the above figure has 1578 digital points, and its polygonal approximation has 77 vertices, so its compression ratio is approximately 20.49.

If $p_u$ and $p_v$ are two consecutive vertices of an approximating polygon, then the departure of digital points $(p_w)$ intervening $p_u$ and $p_v$ $(u < w < v)$ from the side containing the vertices is defined by the absolute perpendicular distance $e_w$ of the points from the line passing through $p_u$ and $p_v$ and is given by

$$e_w = \frac{|(x_w - x_u)(y_v - y_u) - (y_w - y_u)(x_v - x_u)|}{\sqrt{(x_v - x_u)^2 + (y_v - y_u)^2}} \quad (3)$$

The maximum error incurred in approximating the digital points $p_u$ through $p_v$ by a line segment is defined by

$$e_{max} = \max_{u<w<v}(e_w) \quad (4)$$

and the maximum error $(E_{max})$ incurred by a polygonal approximation is defined by

$$E_{max} = \max(e_{max}) \quad (5)$$

which is the maximum of $e_{max}$ over all the sides of the approximating polygon. The sum of the squares of the errors $(E_2)$ is defined as the sum of the squares of the errors $(e_w)$ over all the digital points of a curve, i.e.,

$$E_2 = \sum_{w=1}^{n} {e_w}^2. \quad (6)$$

The approximation shown in Figure 1 generates a maximum error of 2.23, and the sum of the square of the errors is 689.55.

A high compression ratio, a low value of the sum of the square of the errors and a low maximum error are desirable properties of a good approximation. However, as the compression ratio increases, in general, the sum of the squares of the errors increases, and vice versa. This is why the compression ratio and sum of the square of the errors cannot be used to measure the quality of an approximation separately. A similar argument can be made about the relationship between the compression ratio and the maximum error. Because of the conflicting behavior of the compression ratio and error, to assess the quality of an approximation, Sarkar [6] proposed *the figure of merit* ($FoM$), which is defined as the ratio of the compression ratio to the sum of the square of the errors, i.e.,

$$FoM = \frac{CR}{E_2} \quad (7)$$

The higher the value of $FoM$ is, the better the approximation. This measure can be used to compare polygonal approximations (of the same digital curve) with different numbers of vertices, so it can facilitate comparisons among different schemes of polygonal approximation. However, Rosin [7] reported that the two terms (the compression ratio and the sum of the square of the errors) in $FoM$ are not properly balanced. A small change in the compression ratio may result in a large change in the sum of the squares of the errors. Therefore, he introduced fidelity and efficiency of an approximation, defining fidelity as the ratio of the approximation error of an optimal polygon with the same number of vertices as the suboptimal polygon to the approximation error of the suboptimal polygon, expressed as a percentage, viz. $\frac{Error_{optimal}}{Error_{suboptimal}} \times 100$ and efficiency as the ratio of the number of vertices required by the optimal algorithm to produce the same approximation error as the suboptimal polygon to the number of vertices in the suboptimal polygon, expressed in percentage viz. $\frac{m_{optimal}}{m_{suboptimal}} \times 100$. Since it may not always be possible to determine the optimal number of vertices for a specified suboptimal error, interpolation is used to compute the same. He defined the $Merit$ of an approximation as the geometric mean of fidelity and efficiency as in

$$Merit = \sqrt{\frac{Error_{optimal}}{Error_{suboptimal}} \times \frac{m_{optimal}}{m_{suboptimal}}} \times 100. \quad (8)$$

The higher the value of $Merit$ is, the better the approximation in terms of smoothness. The sum of the squares of the errors $E_2$ is usually used to compute the approximation error, and in fact, the above measure involves $E_2$ as the approximation error. Apart from the sum of the squares of the errors, it is also necessary to ensure that the maximum error incurred in an approximation is not too high, especially when the compression ratio is high. This is why, in this communication, in addition to $E_2$, the maximum error $E_{max}$ is also used to measure the merit of an approximation, and this metric is referred to herein as $Merit_{E_{max}}$ and is defined as

$$Merit_{E_{max}} = \sqrt{\frac{(E_{\max})_{optimal}}{(E_{\max})_{suboptimal}} \times \frac{m_{optimal}}{m_{suboptimal}}} \times 100. \quad (9)$$

The higher the value of this measure is, the better the approximation with respect to abnormal distortion. The $Merit$ measure and $Merit_{E_{max}}$ are considered omitting the square root and the factor 100 in the theoretical analysis without loss of generality. The graphs of the measures are drawn (in the Experiments and Statistical Analysis section) after multiplication by a suitable factor for the sake of clarity.

An optimal algorithm has its inherent drawback in that it results in approximation with highly undesirable distortion, especially when the number of vertices is significantly low. More importantly, the running time of an optimal algorithm, especially for large curves with many vertices, is significantly high. The last factor leads to a significantly large amount of time involved in testing the goodness of fit of a suboptimal technique via Rosin's $Merit$ (8) and $Merit_{E_{max}}$ (9).

Following the deficiency of Sarkar's *figure of merit* and *Rosin's Merit* measure (which will henceforth be called Rosin's measure for the sake of convenience), researchers started using the reciprocal of the figure of merit and other measures derived from the sum of the square of the errors (or maximum error) and compression ratio instead of Rosin's measure. These measures are defined by

$$WE_2 = \frac{E_2}{CR^2} \quad (10)$$

$$WE_3 = \frac{E_2}{CR^3} \quad (11)$$

$$WE_\infty = \frac{E_{max}}{CR}. \quad (12)$$

The last measure (12) indicates the presence/absence of excessive distortion in the approximation. The smaller the values of these measures are, the better the approximation. The measures $WE_2$ and $WE_3$ intuitively indicate the degree of smoothness of an approximation, and $WE_\infty$ intuitively ensures that a high compression ratio does not result in a highly distorted approximation. A low value of $WE_2$ and $WE_3$ is supposedly indicative of a smooth approximation with a relatively reasonable number of vertices, and a low value of $WE_\infty$ is supposedly indicative of an approximation that is not distorted and intuitively has a reasonable number of vertices.

In this communication, an attempt is made to investigate whether weighted figure merit is related to Rosin's measure in assessing the merit of a suboptimal approximation and, in this way, the reciprocal of $FoM$ defined by

$$WE = \frac{E_2}{CR}. \tag{13}$$

is also investigated for a possible relationship with Rosin's measure. This measure is also referred to as a weighted figure of merit in this communication. As shown in the subsequent theoretical analysis, experimental studies and statistical analysis, the measures referred to as the weighted figure of merit and Rosin's measure are independent of each other; hence, it is not justifiable to use a weighted figure of merit instead of Rosin's measure to compare among suboptimal schemes of polygonal approximation.

The theoretical analysis is presented in the next section (Section III) to explore the relationship between the weighted figure of merit $WE_\nu$ for ν = 1, 2, and 3, including $WE_\infty$ and Rosin's $Merit$ measure and the measure $Merit_{E_{max}}$. An overview of some of the polygonal approximation schemes is presented in Section IV to shed light on the current literature on various schemes of polygonal approximation. The results of experiments in support of theoretical analysis using various schemes of polygonal approximation are presented and analyzed in Section V. This section also presents a statistical analysis to explore the possibility of a relationship between the weighted figure of merit and Rosin's measure and $Merit_{E_{max}}$. Finally, in Section VI, it is concluded that the weighted figure of merit cannot replace Rosin's measure and $Merit_{E_{max}}$ to assess the quality of polygonal approximations produced by suboptimal schemes.

## III THEORETICAL ANALYSIS

The following theorems establish that Rosin's measure and weighted figure of merit ($WE, WE_2, WE_3$ and $WE_\infty$) are independent of each other. The proof of the theorems is based on intuition.

**Theorem I**

*The Rosin's measures $Merit$ and $WE_{suboptimal}$ are independent of each other.*

*Proof:*

Rosin's *Merit* measure (omitting the square root and percentage factor for the sake of convenience but without loss of precision and generality) can be written as $\frac{Error_{optimal}}{Error_{suboptimal}} \times \frac{m_{optimal}}{m_{suboptimal}}$

$$= \frac{Error_{optimal}}{Error_{suboptimal}} \times \frac{\frac{m_{optimal}}{n}}{\frac{m_{suboptimal}}{n}} = \frac{Error_{optimal}}{\frac{n}{m_{optimal}}} \times \frac{\frac{n}{m_{suboptimal}}}{Error_{suboptimal}} = \frac{Error_{optimal}}{\frac{n}{m_{optimal}}} \times \frac{1}{\frac{Error_{suboptimal}}{\frac{n}{m_{suboptimal}}}} = \frac{Error_{optimal}}{CR_{optimal}} \times \frac{1}{\frac{Error_{suboptimal}}{CR_{suboptimal}}} = \frac{WE_{optimal}}{WE_{suboptimal}}.$$

Since $\frac{WE_{optimal}}{WE_{suboptimal}}$, a simplified version of Rosin's *Merit* measure, depends on $WE_{optimal}$ as well as $WE_{suboptimal}$ and $WE_{optimal}$ is not a constant, it depends not only on the compression ratio ($CR$) but also on the error value, and these two measures have conflicting behavior; hence, one cannot conclude that Rosin's measure is related to $WE_{suboptimal}$ only. Following the same line of argument, one can conclude that $WE_{suboptimal}$ is not related to Rosin's measure.

**Theorem II**

*It is not possible to derive a theoretical relationship between Rosin's measure and $(WE_2)_{suboptimal}$.*

*Proof:*

The Rosin's *Merit* measure after the square root and percentage factor are omitted is

$$\frac{Error_{optimal}}{Error_{suboptimal}} \times \frac{m_{optimal}}{m_{suboptimal}}$$

$$= \frac{Error_{optimal}}{Error_{suboptimal}} \times \frac{m_{optimal}}{m_{suboptimal}} \times \frac{m_{optimal}}{m_{suboptimal}} \times \frac{m_{suboptimal}}{m_{optimal}}$$

$$= \frac{Error_{optimal}}{Error_{suboptimal}} \times \frac{\frac{m_{optimal}}{n}}{\frac{m_{suboptimal}}{n}} \times \frac{\frac{m_{optimal}}{n}}{\frac{m_{suboptimal}}{n}} \times \frac{\frac{m_{suboptimal}}{n}}{\frac{m_{optimal}}{n}}$$

$$= \frac{Error_{optimal}}{Error_{suboptimal}} \times \frac{\frac{n}{m_{suboptimal}}}{\frac{n}{m_{optimal}}} \times \frac{\frac{n}{m_{suboptimal}}}{\frac{n}{m_{optimal}}} \times \frac{\frac{n}{m_{optimal}}}{\frac{n}{m_{suboptimal}}}$$

$$= \frac{Error_{optimal}}{Error_{suboptimal}} \times \frac{CR_{suboptimal}}{CR_{optimal}} \times \frac{CR_{suboptimal}}{CR_{optimal}} \times \frac{CR_{optimal}}{CR_{suboptimal}}$$

$$= \frac{Error_{optimal}}{(CR_{optimal})^2} \times \frac{1}{\frac{Error_{suboptimal}}{(CR_{suboptimal})^2}} \times \frac{CR_{optimal}}{CR_{suboptimal}}$$

$$= \frac{(WE_2)_{optimal}}{(WE_2)_{suboptimal}} \times \frac{CR_{optimal}}{CR_{suboptimal}}$$

The expression $\frac{(WE_2)_{optimal}}{(WE_2)_{suboptimal}} \times \frac{CR_{optimal}}{CR_{suboptimal}}$ is another form of Rosin's *Merit* measure (after the square root and the numerical factor are omitted), and it may be observed that it depends not only on $(WE_2)_{optimal}$ but also on $(WE_2)_{suboptimal}$, $CR_{optimal}$ and $CR_{suboptimal}$. It is not possible to assume that $(WE_2)_{optimal}$ is constant because it depends not only on the error value but also on the compression ratio, and these two measures have conflicting behavior. Moreover, neither $CR_{optimal}$ nor $CR_{suboptimal}$ are constants; rather, their values vary from approximation to approximation. Hence, one cannot conclude that Rosin's *Merit* measure is related to $(WE_2)_{suboptimal}$ only. Following the same line of argument, one can conclude that $(WE_2)_{suboptimal}$ is not related to Rosin's measure.

**Theorem III**

*Rosin's measure and* $(WE_3)_{suboptimal}$ *are independent of each other.*

*Proof:*

The proof is similar to that of theorem II.

**Theorem IV**

*The measures* $Merit_{E_{max}}$ *and* $(WE_\infty)_{suboptimal}$ *are independent of each other.*

*Proof:*

The measure $Merit_{E_{max}}$ (omitting the square root and percentage factor for the sake of convenience but without loss of precision and generality) can be written as

$$\frac{(E_{\max})_{optimal}}{(E_{\max})_{suboptimal}} \times \frac{m_{optimal}}{m_{suboptimal}} = \frac{(E_{\max})_{optimal}}{(E_{\max})_{suboptimal}} \times \frac{\frac{m_{optimal}}{n}}{\frac{m_{suboptimal}}{n}}$$

$$= \frac{(E_{\max})_{optimal}}{\frac{n}{m_{optimal}}} \times \frac{\frac{n}{m_{suboptimal}}}{(E_{\max})_{suboptimal}}$$

$$= \frac{(E_{\max})_{optimal}}{CR_{optimal}} \times \frac{CR_{suboptimal}}{(E_{\max})_{suboptimal}}$$

$$= \frac{(WE_\infty)_{optimal}}{(WE_\infty)_{suboptimal}}$$

The value of $\frac{(WE_\infty)_{optimal}}{(WE_\infty)_{suboptimal}}$, a simplified version of $Merit_{E_{max}}$ (omitting the square root and the factor 100), depends on $(WE_\infty)_{optimal}$ as well as $(WE_\infty)_{suboptimal}$ and $(WE_\infty)_{optimal}$ is not a constant because it depends not only on the compression ratio ($CR_{optimal}$) but also on the error value $E_{\max}$, and these two measures have conflicting behavior; hence, one cannot conclude that measuring $Merit_{E_{max}}$ is related to $(WE_\infty)_{suboptimal}$ only. Following the same line of argument, one can conclude that $(WE_\infty)_{suboptimal}$ is not related to the $Merit_{E_{max}}$ measure.

## IV SOME POLYGONAL APPROXIMATION SCHEMES

In this section, an overview of some of the schemes for polygonal approximation is presented, and these schemes are then used to validate the theoretical analysis presented in the last section.

There are several schemes for polygonal approximation, many of which are supervised (parametric), and there are other schemes that are unsupervised (nonparametric) in nature. Researchers have also developed a framework (e.g., [8]) that facilitates the conversion of a parametric scheme into a nonparametric scheme. Among the former schemes are iterative splitting ([9], [10]), iterative split-and-merge (e.g., [11]), sequential (([12], [13], [14]) and iterative point elimination, which may be considered an iterative merging scheme ([15], [16], [17], [18], [19]). All these schemes can be converted into nonparametric versions via a suitable framework. There also exists iterative point elimination, which is nonparametric in nature, and there are other nonparametric approaches that are hybrid in nature because by nature, they are a mix of conventional approaches (split, merge, sequential) and iterative point elimination.

Fernandez-Gracia et al. [20] proposed an unsupervised scheme as an improvement over the symmetric versions [21] of Ramer [9] and Doughlas-Pecker [10], which [21] were also unsupervised in nature. The latter scheme [21] determines two points on a curve as initial points—one of the initial points is the one that is at the farthest distance from the centroid of a curve, and the other is at the maximum distance from the point already determined. The segments obtained are then subjected to iterative subdivision at a point most distant from the segment, taking into account the symmetry in the distribution of the vertices. These vertices are called non-initial points and are assigned a *significance value* defined by the absolute perpendicular distance of the point from the line segment, which is used to detect the most distant point. If the maximum of the significance value of the non-initial points is zero, then the initial points are assigned a significance value of unity; otherwise, the significance value of the initial points is the maximum of the maximum significance value of the non-initial points and the largest distance on the curve boundary from its centroid. A normalized significance curve is considered to determine a threshold automatically, and the threshold is used to detect the vertices of the approximation. Four different methods—proximity, distance, Rosin and adaptive—are used to determine the threshold. The adaptive threshold method produces the best result except in some exceptional situations where the use of the proximity method is recommended. As an improvement of this work, Fernandez-Gracia et al. [21] use a convex hull to determine the initial points, use an adaptive threshold on the normalized significance curve and subject the resulting approximation to refinement through elimination of pseudo vertices and subsequent vertex adjustment. The last two works are similar, with the latter improving the performance of the former. Another unsupervised scheme with appreciable quality of approximation is that of Madrid-Cuevas et al. [22]. Here, convex hull decomposition of the input curve is used, and the Prasad et al. [8] framework is used for further decomposition without using any input parameter (threshold). Convex hull decomposition generates too many vertices, especially in the circular region of a curve, apart from the noisy convex points. Moreover, convex hull decomposition does not capture concave turning, which is why the Prasad et al. framework is used to pick up more vertices, some of which may be pseudo. In an attempt to eliminate pseudo vertices and to produce an aesthetic approximation, a subsequent four-vertex merging scheme is used through minimization of the weighted figure of merit $WE_2$ to remove noisy vertices retaining the unsupervised nature of the scheme. This scheme, although slightly involved in the execution process, produces good approximations, as revealed by Rosin's measure and visual inspection. Parvez and Mahmoud [12] proposed another unsupervised scheme wherein they obtained the most important vertices (that persist through scales), called cutoff points, and then applied unsupervised decomposition of the consecutive segments to minimize the weighted figure of merit. The cutoff points are high curvature points determined through an iterative constrained collinear-point suppression technique. The strength of the break points is computed, and the curve is then sorted first with respect to strength and then with respect to the distance of the break points from the centroid of the curve. The break points are eliminated one after another starting with the weakest break point, and every time a break point is eliminated from the prospective set of vertices of the polygonal approximation, the strength of the vertices is adjusted. The constrained collinear-point suppression method deletes a break point (pseudo vertex) if its perpendicular distance from the line segment joining its adjacent break points is less than a threshold and its adjoining segment is also farther from it by more than the threshold. Constrained collinear-point suppression is used to ensure that sharp points are retained and that self-intersections are not created through the suppression process. The iterative process starts with a threshold of 0.5, is incremented with a step size of 0.5 and is terminated when two successive iterations produce the same number of vertices. The segments defined by a pair of consecutive cutoff points are then refined to generate intermediate vertices through local optimization of any of the weighted figures of merit $WE$, $WE_2$ and $WE_3$ over the segment joining the adjacent cutoff points. The key takeaway from this scheme is the concept of cut-points and the independence of the final approximation of the choice of weighted figure of merit. In contrast to Madrid-Cuevas et al., Parvez and Mahmoud used local minimization of the weighted figure of merit. Madrid-Cuevas et al. used two phases of vertex insertion followed by merging, whereas Parvez and Mahmoud used the coarsest possible approximation defined by the cutoff points and then refined it through the necessary number of vertex insertions. Parvez [24] proposed another automatic linear approximation of digital curves reusing constrained-collinear-point suppression, as in [23], and then either relocating vertices within a neighborhood or deleting vertices through optimization of an error measure. The vertices are not relocated in any of the positions between adjacent vertices, as it is in Masood's stabilization scheme [25]; rather, they are relocated to a point within the neighborhood of a vertex. The neighborhood of a vertex is determined during iterative constrained collinear-point suppression. If the relocation error is found to be greater than the deletion error, then the vertex is deleted; otherwise, the vertex is relocated. The improvement in the approximation because of vertex relocation may not be significant because there is a narrow permissible region for vertex movement, which is not the case in Masood's stabilization scheme [25]. The vertex with the least strength is selected first for relocation/deletion. The output vertices are not necessarily on the boundary of the input curve.

## V EXPERIMENTS AND STATISTICAL ANALYSIS

Four algorithms are used here to explore the possibility of a relationship between Rosin's measure and the weighted figure of merit: $WE_{suboptimal}$, $(WE_2)_{suboptimal}$, $(WE_3)_{suboptimal}$ and $(WE_\infty)_{suboptimal}$. The first three measures are compared with Rosin's $Merit$ measure, and the last one is compared with $Merit_{E_{max}}$ via the same algorithms. The algorithms used for comparison are Madrid-Cuevas et al. [22], Fernandez-Gracia et al. [20], Masood's stabilized scheme [25] and Masood's [15] scheme. There are other iterative point elimination schemes ([16], [17], [18], [19]) that use the same principle as Masood's scheme, but the latter is found to produce better approximations than the former.

The approaches of Madrid-Cuevas et al. [22] and Fernandez Gracia et al. [20] are unsupervised in nature, so a user's intervention is not required to specify either the number of vertices or a threshold on the error value. However, Masood's iterative point elimination and Masood's stabilized scheme require user intervention. This is why the experiments are carried out through the generation of polygonal approximations via the technique of Madrid-Cuevas et al., and the number of vertices of these approximations is used to generate polygonal approximations via Masood's algorithm and Masood's stabilized algorithm. The algorithm of Madrid-Cuevas et al. is selected for this purpose instead of that of Fernandez-Gracia et al. [20] because the former is found to produce more aesthetic approximations than the latter.

Rosin's $Merit$ measure and $Merit_{E_{max}}$ are computed for Madrid-Cuevas et al. via an approximate version of Perez and Vidal's [26] optimal scheme to reduce the execution time of the original scheme. Three iterations of the Perez and Vidal schemes are performed in the approximate version to reduce the time required for comparison, as in [27]. The first iteration is used to determine the starting point for the algorithm, which is used as the starting point in the subsequent two iterations. The second vertex generated by the Perez and Vidal algorithms using the number of vertices of the suboptimal approximation as input is taken as the starting point. The sum of the square of the errors generated by the optimal algorithm is computed as $Error_{optimal}$ for the number of vertices ($m_{suboptimal}$) generated by the suboptimal algorithm using the starting point obtained from the first iteration. The third iteration of the optimal algorithm is carried out with the same starting point and is used to interpolate the number of vertices ($m_{optimal}$) that would be generated by the optimal algorithm for the sum of the square of the errors ($Error_{suboptimal}$) produced by the suboptimal algorithm. Rosin's $Merit$ measure is then computed using the errors and the number of vertices thus obtained. The measure $Merit_{E_{max}}$ requires $(E_{\max})_{optimal}$ corresponding to $m_{optimal}$ and $m_{suboptimal}$ corresponding to $(E_{\max})_{suboptimal}$ and can be computed in a similar way as described for the case of the sum of the square of the errors. The measures, viz. $WE_{suboptimal}$, $(WE_2)_{suboptimal}$, $(WE_3)_{suboptimal}$ and $(WE_\infty)_{suboptimal}$ are computed via the suboptimal algorithm and are compared with Rosin's measure and $Merit_{E_{max}}$. The first three measures are compared with $Merit$, and the fourth one is compared with $Merit_{E_{max}}$. The images from the MPEG7 dataset [28] are used for comparison.

The higher the measures $Merit$ and $Merit_{E_{max}}$ are, the better the approximation is, and the lower the weighted figure of merit is, the better the approximation is; hence, the measures $Merit$ and $Merit_{E_{max}}$ are compared with the reciprocal of the weighted figure of merit viz. $WE_{suboptimal}$, $(WE_2)_{suboptimal}$ and $(WE_\infty)_{suboptimal}$. The original value of $(WE_3)_{suboptimal}$, instead of its reciprocal, is used in comparison for the reasons mentioned later. The $x$-axis, in the graphical investigation of the relationship between the two measures, shows different digital curves from the MPEG7 dataset, and the $y$-axis indicates the measure $Merit/Merit_{E_{max}}$ and the weighted figure of merit in two different diagrams. The latter are plotted as points on the 2D plane for each curve, and the points are joined in the sequence of the curves via a straight line segment, leading to a line diagram. The line diagram for $Merit/Merit_{E_{max}}$ is shown in yellow, and the line diagram for the weighted figure of merit is drawn in blue. These diagrams show how the measure $Merit/Merit_{E_{max}}$ and weighted figure of merit change for different curves. It facilitates the investigation of whether the peaks and valleys and the rise and fall in the line diagram produced by $Merit/Merit_{E_{max}}$ match those in the line diagram for the reciprocal of $WE_{suboptimal}$, $(WE_2)_{suboptimal}$ and $(WE_\infty)_{suboptimal}$. The comparison of $(WE_3)_{suboptimal}$ with the $Merit$ measure is treated in a slightly different way in that the peaks and valleys (rise and fall) of $(WE_3)_{suboptimal}$ are compared with the valleys and peaks (fall and rise) of the $Merit$ measure. The peaks and valleys (rise and fall) of the $Merit$ measure are expected to match the valleys and peaks (fall and rise), respectively, of $(WE_3)_{suboptimal}$ if the measures are related. If the peaks and valleys of the $Merit$ and $Merit_{E_{max}}$ measures match those of the weighted figure of merit and the rise and fall of the $Merit$ and $Merit_{E_{max}}$ measures dictate a rise and fall, respectively, in the reciprocal of the weighted figure of merit, $WE_{suboptimal}$, $(WE_2)_{suboptimal}$ and $(WE_\infty)_{suboptimal}$, then it can be concluded that the two measures behave in a similar manner ($(WE_3)_{suboptimal}$ is treated in a different way), so they are related. However, as discovered subsequently, there is no reason to conclude that the two measures are related.

Figure 2 shows the graphical representation of $Merit$ (Rosin's measure) in yellow and the reciprocal of $WE_{suboptimal}$ in blue in the form of a line diagram for the Madrid--Cuevas et al. scheme. The values of Rosin's measure and the reciprocal of $WE_{suboptimal}$ are scaled by a suitable factor to provide clarity in the line diagrams. The scaling, however, does not affect the valleys (minima points) and peaks (maxima points) or the events of rise and fall in the line diagram.

The Figure 2 shows that although there are similarities in the behavior of the two line diagrams, there is also a difference between the two. The peaks and valleys on the yellow line diagram (Rosin's measure) do not always match those on the blue line diagram. Moreover, a rise (fall) in the yellow line diagram does not dictate a rise (fall) in the blue line diagram. This finding verifies the theoretical finding that Rosin's measure and the reciprocal of $WE_{suboptimal}$ are independent of each other. To facilitate comprehension of this observation in detail, the graph of the line diagrams is further annotated with additional vertical lines drawn from the horizontal axis through the corresponding points on the line diagrams. The vertical lines are drawn in yellow at the $i^{th}$ curve along $x-axis$ if the two line diagrams simultaneously increase or decrease as one moves from the $i^{th}$ curve to the $(\mathrm{i}+1)^{st}$ curve; otherwise, a blue line is drawn. The presence of a mix of yellow and blue vertical lines in Figure 2 is indicative of the independence of $Merit$ and $WE_{suboptimal}$.

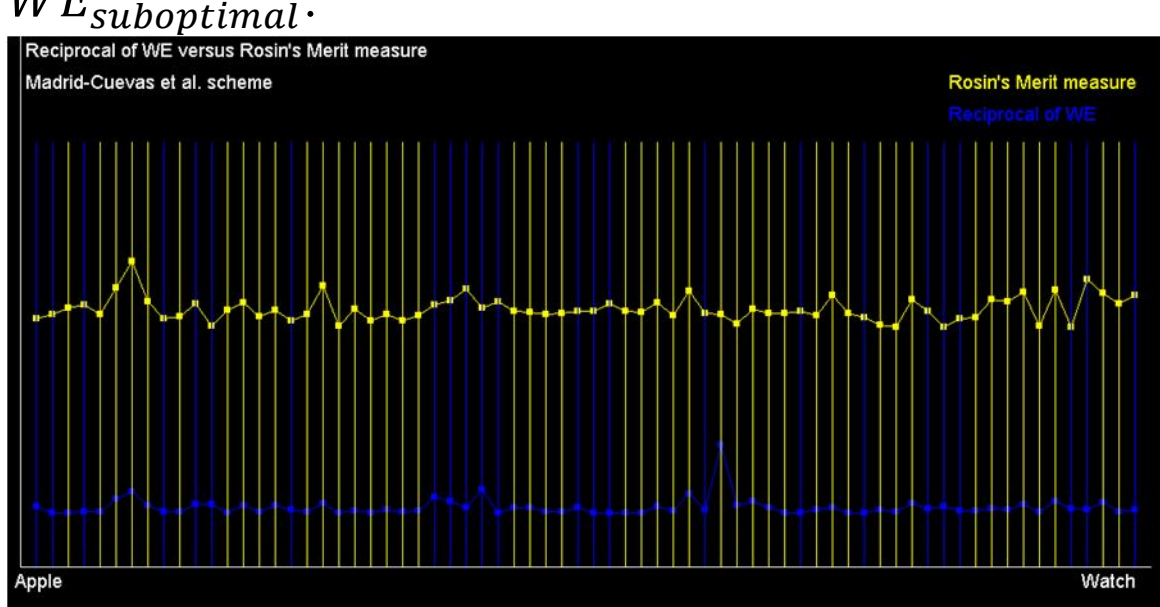

**FIGURE 2 A plot of the reciprocal of $WE_{suboptimal}$ (in blue) and Rosin's $Merit$ measure (in yellow) via the Madrid--Cuevas et al. scheme. The graph is annotated with vertical lines shown in yellow and blue to facilitate comparison of the two line diagrams. The yellow lines indicate similarity in the pattern of the line diagram, and the blue lines indicate dissimilarity. As there is a mix of yellow and blue lines, the two measures are independent of each other.**

Figure 3 shows the line diagram for Rosin's measure and the reciprocal of $(WE_2)_{suboptimal}$ suitably scaled to provide clarity in line diagrams via the Madrid--Cuevas et al. scheme. The line diagram is annotated with additional vertical lines to facilitate the comparison of the behavior of the measures. The two line diagrams, along with the annotated vertical lines, show that the valleys and peaks on the line diagram for Rosin's measure do not necessarily match those on the line diagram of the reciprocal of $(WE_2)_{suboptimal}$ and that a rise (fall) on the line diagram does not necessarily dictate the same on the line diagram of the reciprocal of $(WE_2)_{suboptimal}$.

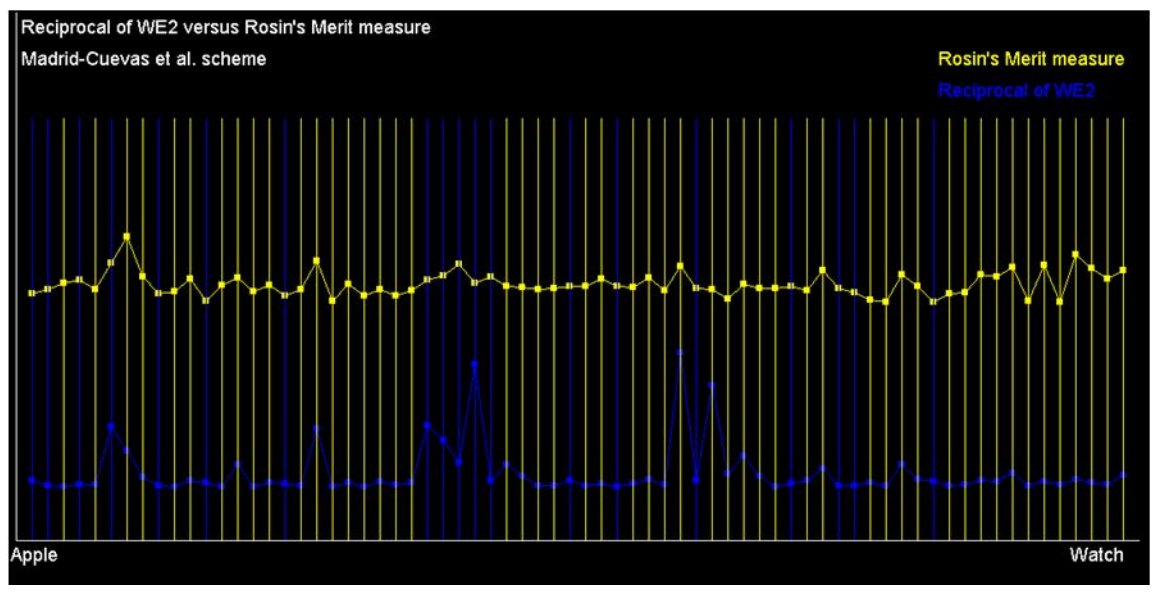

**FIGURE 3 A plot of the reciprocal of $(WE_2)_{suboptimal}$ (in blue) and Rosin's $Merit$ measure (in yellow) via the Madrid--Cuevas et al. scheme. The graph is annotated with vertical lines shown in yellow and blue to facilitate comparison of the two line diagrams. The yellow lines indicate similarity in the pattern of the line diagram, and the blue lines indicate dissimilarity. As there is a mix of yellow and blue lines, the two measures are independent of each other.**

This figure again shows enough evidence to claim that the two measures, viz. $Merit$ and $(WE_2)_{suboptimal}$ are not related.

When Rosin's measure is compared with the reciprocal of $(WE_3)_{suboptimal}$, the value of the reciprocal is significantly high because the value of $(WE_3)_{suboptimal}$ is significantly small because of the presence of the third power of the compression ratio in the denominator of $WE_3$. Therefore, it is proposed to consider $(WE_3)_{suboptimal}$ itself (instead of its reciprocal) to draw its line diagram after scaling it by a suitable factor.

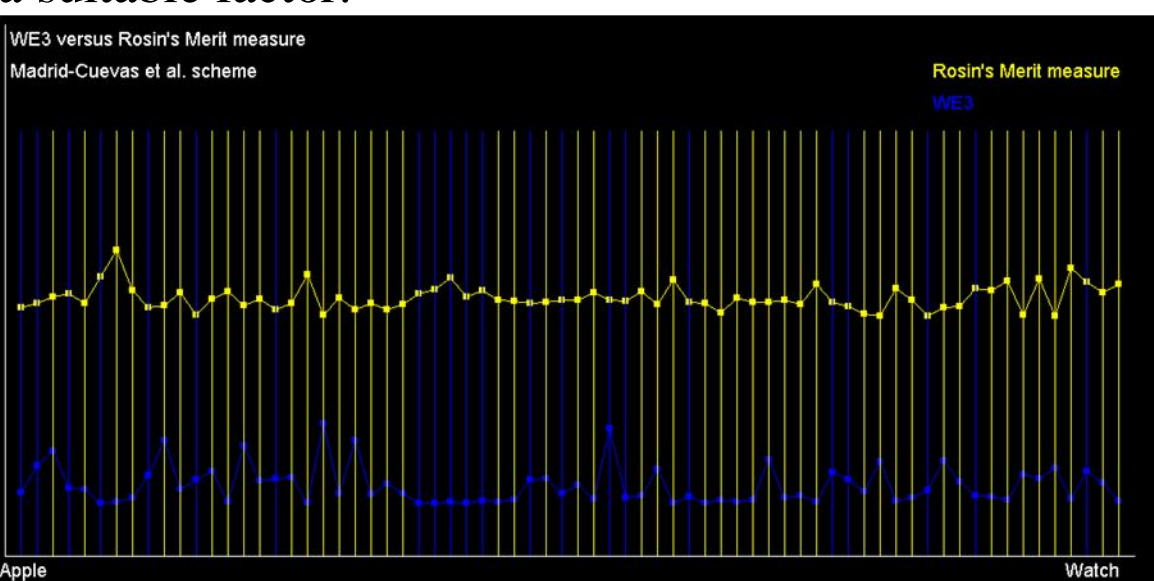

**FIGURE 4 A plot of $(WE_3)_{suboptimal}$ (in blue) and Rosin's $Merit$ measure (in yellow) via the Madrid--Cuevas et al. scheme. The graph is annotated with vertical lines shown in yellow and blue to facilitate comparison of the two line diagrams. The yellow lines indicate similarity in the pattern of the line diagram, and the blue lines indicate dissimilarity. As there is a mix of yellow and blue lines, the two measures are independent of each other.**

Figure 4 shows the line diagram for Rosin's measure in yellow and $(WE_3)_{suboptimal}$ in blue. It may be observed from the line diagrams that a peak or a valley on the line diagram for Rosin's measure does not dictate a valley or a peak, respectively, on the line diagram for $(WE_3)_{suboptimal}$, and a rise or a fall on the line diagram for Rosin's measure does not dictate a fall or a rise, respectively, on the line diagram for $(WE_3)_{suboptimal}$. The figure shown earlier is annotated with vertical lines to facilitate a comparison of the behavior of the two line diagrams.

Figure 5 shows the line diagram (in yellow) for the $Merit_{E_{max}}$ measure and the reciprocal of $(WE_\infty)_{suboptimal}$ (blue line diagram). It may be observed from the line diagrams with the help of the annotated vertical lines in yellow and blue that they do not always match each other with respect to peaks and valleys and rise and fall on the line diagrams, as manifested by a mix of yellow and blue vertical lines. This shows that the two measures are independent of each other.

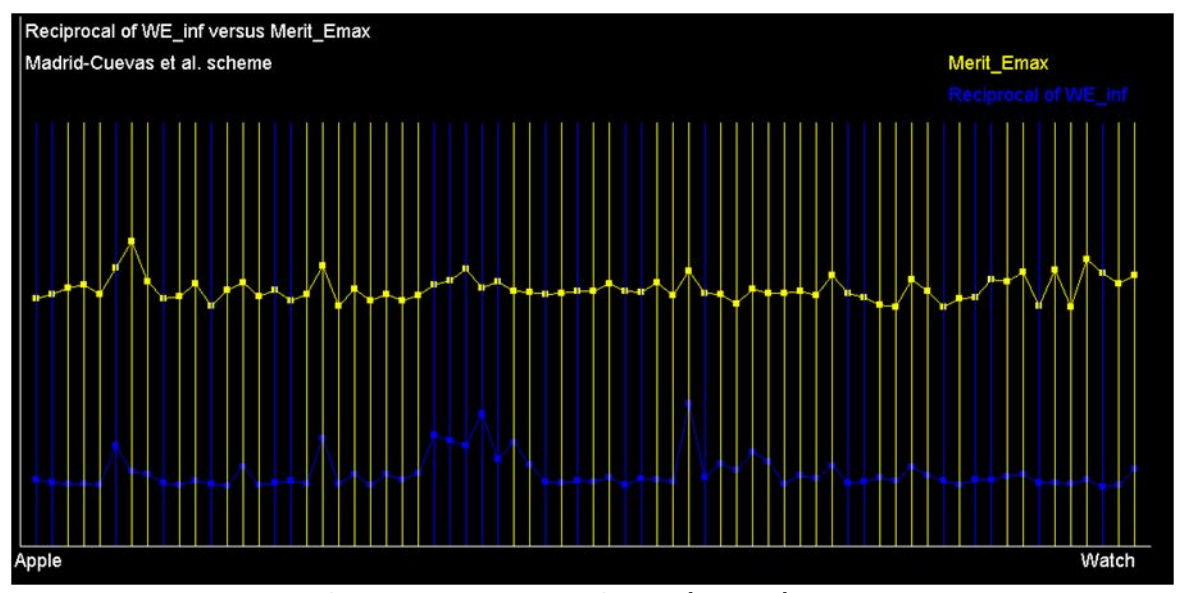

**FIGURE 5** **A plot of the reciprocal of the $(WE_\infty)_{suboptimal}$ (in blue) and $Merit_{E_{max}}$ measures (in yellow) via the Madrid--Cuevas et al. scheme. The graph is annotated with vertical lines shown in yellow and blue to facilitate comparison of the two line diagrams. The yellow lines indicate similarity in the pattern of the line diagram, and the blue lines indicate dissimilarity. As there is a mix of yellow and blue lines, the two measures are independent of each other.**

Figures 6, 7, 8 and 9 show graphs similar to those in Figures 1 through 4 for the Fernández García et al. [10] scheme. Figures 10, 11, 12 and 13 show graphs similar to those in Figures 1 through 4 for Masood's stabilized scheme [5], and Figures 14, 15, 16 and 17 show graphs similar to those in Figures 1 through 4 for Masood's scheme [4]. These figures indicate that the behavior of these line diagrams is similar to that of lines 1 through 4. Therefore, Rosin's measure and the weighted figure of merit are independent of each other, which is why measuring the weighted figure of merit does not furnish information about Rosin's measure in the assessment of a polygonal approximation scheme.

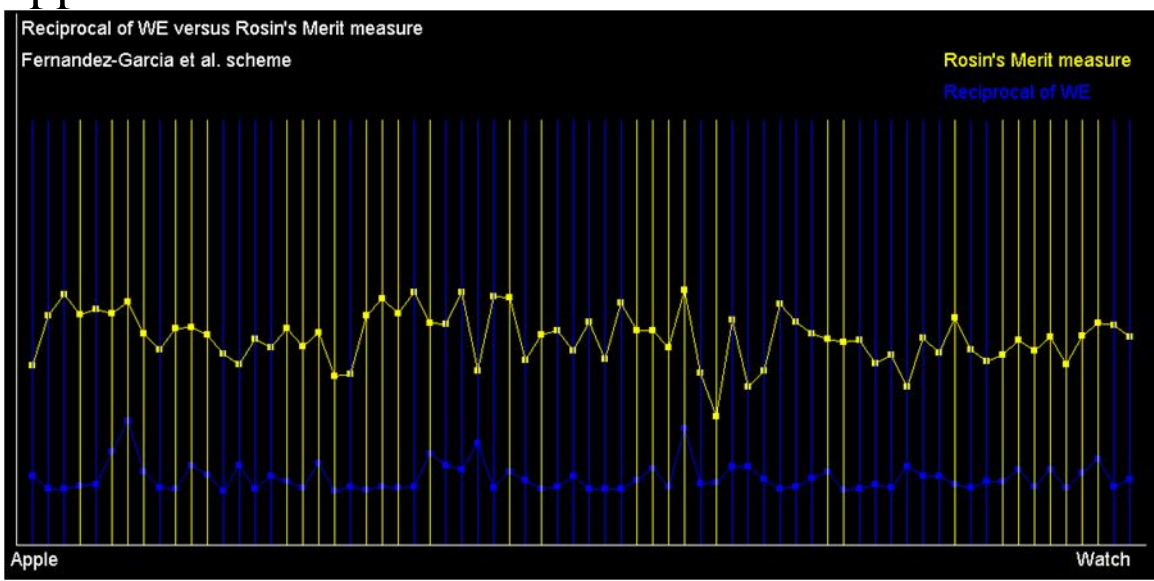

**FIGURE 6** **A plot of the reciprocal of $WE_{suboptimal}$ (in blue) and Rosin's $Merit$ measure (in yellow) via the Fernandez–Garcia et al. scheme. The graph is annotated with vertical lines shown in yellow and blue to facilitate comparison of the two line diagrams. The yellow lines indicate similarity in the pattern of the line diagram, and the blue lines indicate dissimilarity. As there is a mix of yellow and blue lines, the two measures are independent of each other.**

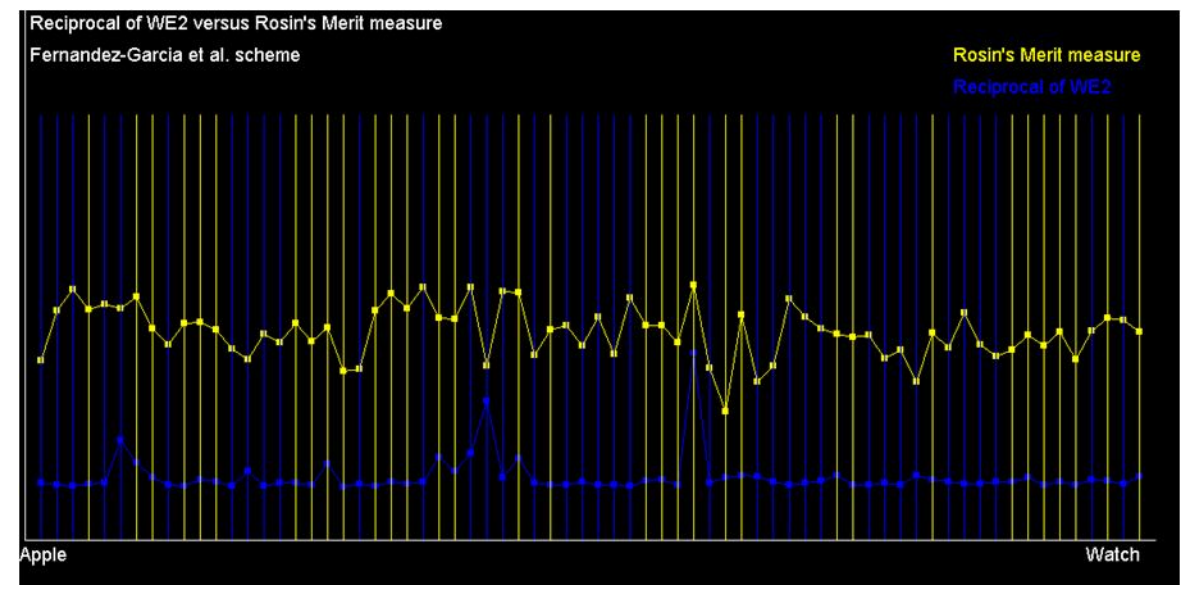

**FIGURE 7** **A plot of the reciprocal of $(WE_2)_{suboptimal}$ (in blue) and Rosin's $Merit$ measure (in yellow) via the Fernández García et al. scheme. The graph is annotated with vertical lines shown in yellow and blue to facilitate comparison of the two line diagrams. The yellow lines indicate similarity in the pattern of the line diagram, and the blue lines indicate dissimilarity. As there is a mix of yellow and blue lines, the two measures are independent of each other.**

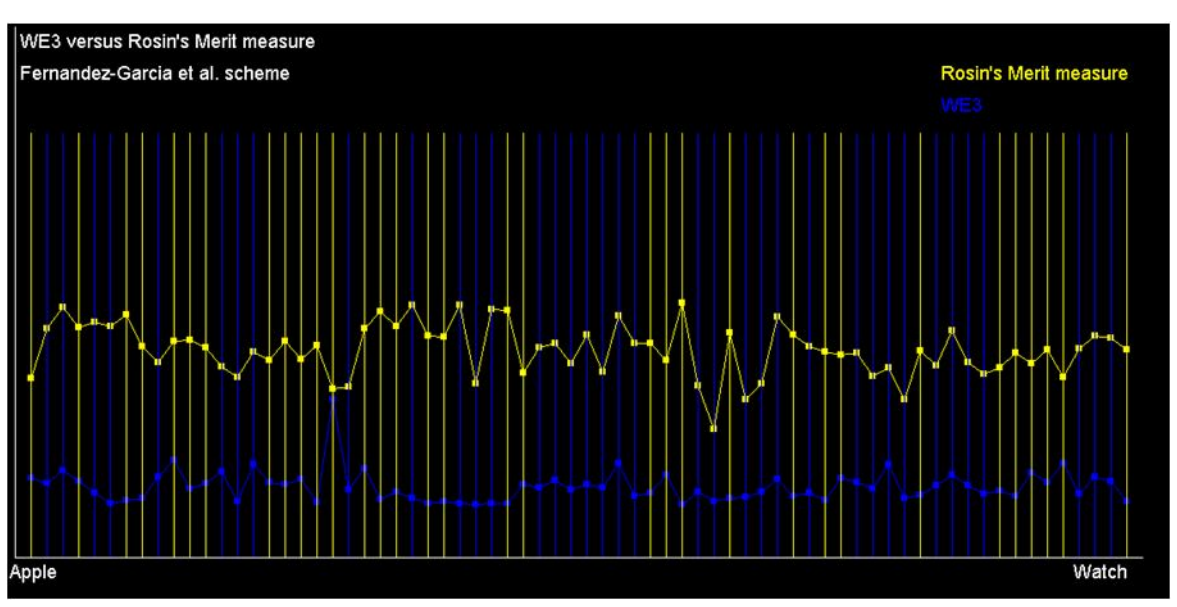

**FIGURE 8** **A plot of $(WE_3)_{suboptimal}$ (in blue) and Rosin's $Merit$ measure (in yellow) via the Fernández García et al. scheme. The graph is annotated with vertical lines shown in yellow and blue to facilitate comparison of the two line diagrams. The yellow lines indicate similarity in the pattern of the line diagram, and the blue lines indicate dissimilarity. As there is a mix of yellow and blue lines, the two measures are independent of each other.**

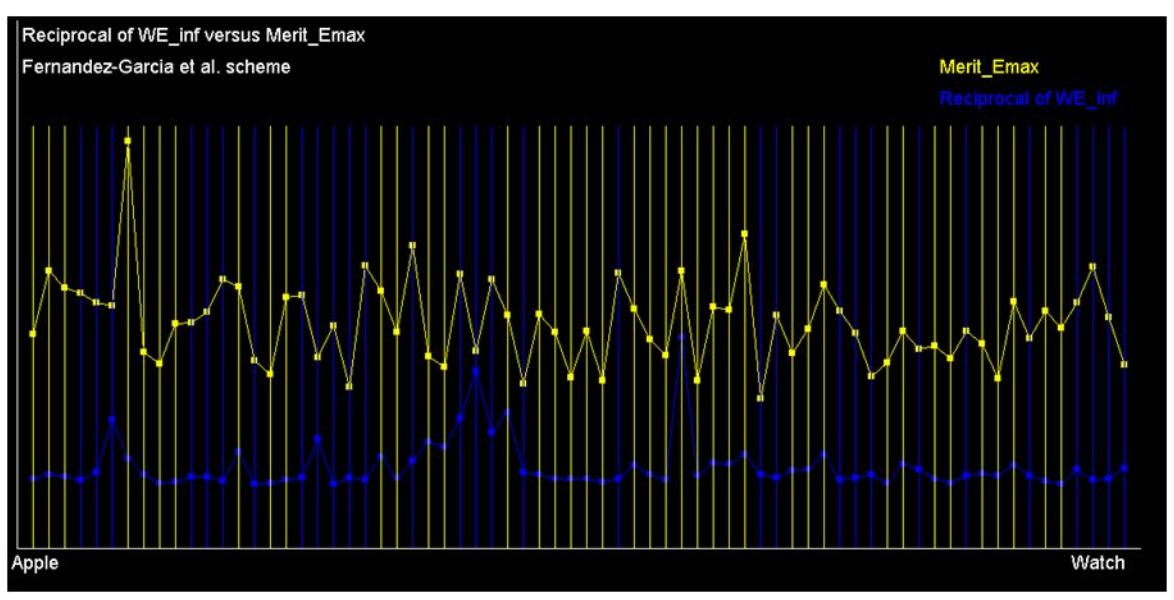

**FIGURE 9** **A plot of the reciprocal of $(WE_\infty)_{suboptimal}$ (in blue) and the measure $Merit_{E_{max}}$ (in yellow) via the Fernández--García et al. scheme. The graph is annotated with vertical lines shown in yellow and blue to facilitate comparison of the two line diagrams. The yellow lines indicate similarity in the pattern of the line diagram, and the blue lines indicate dissimilarity. As there is a mix of yellow and blue lines, the two measures are independent of each other.**

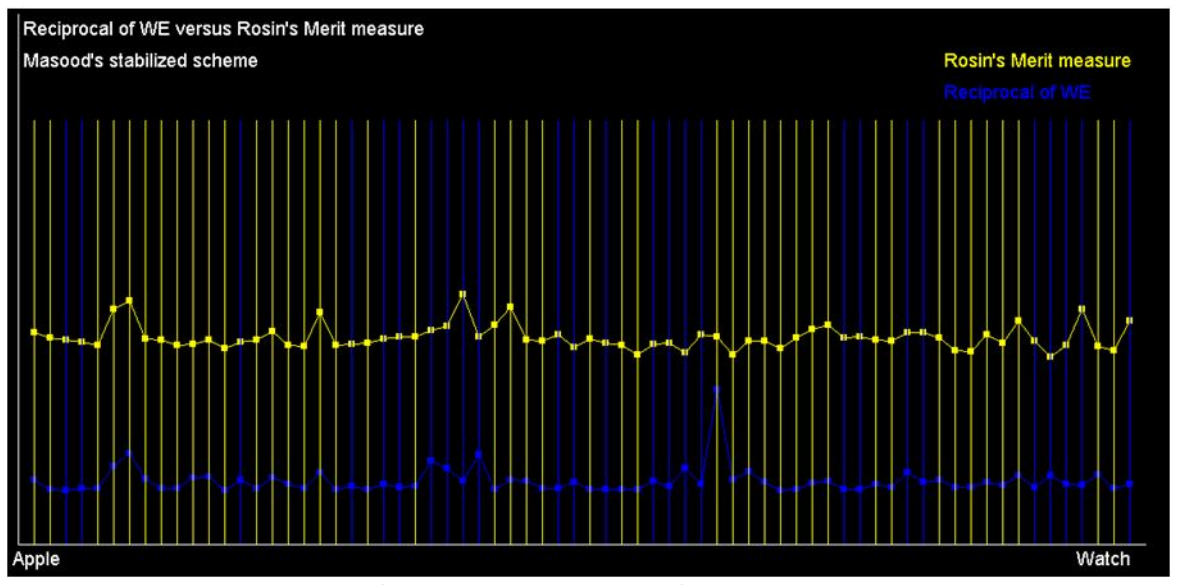

**FIGURE 10** **A plot of the reciprocal of $WE_{suboptimal}$ (in blue) and Rosin's $Merit$ measure (in yellow) using Masood's stabilized scheme.**

The graph is annotated with vertical lines shown in yellow and blue to facilitate comparison of the two line diagrams. The yellow lines indicate similarity in the pattern of the line diagram, and the blue lines indicate dissimilarity. As there is a mix of yellow and blue lines, the two measures are independent of each other.

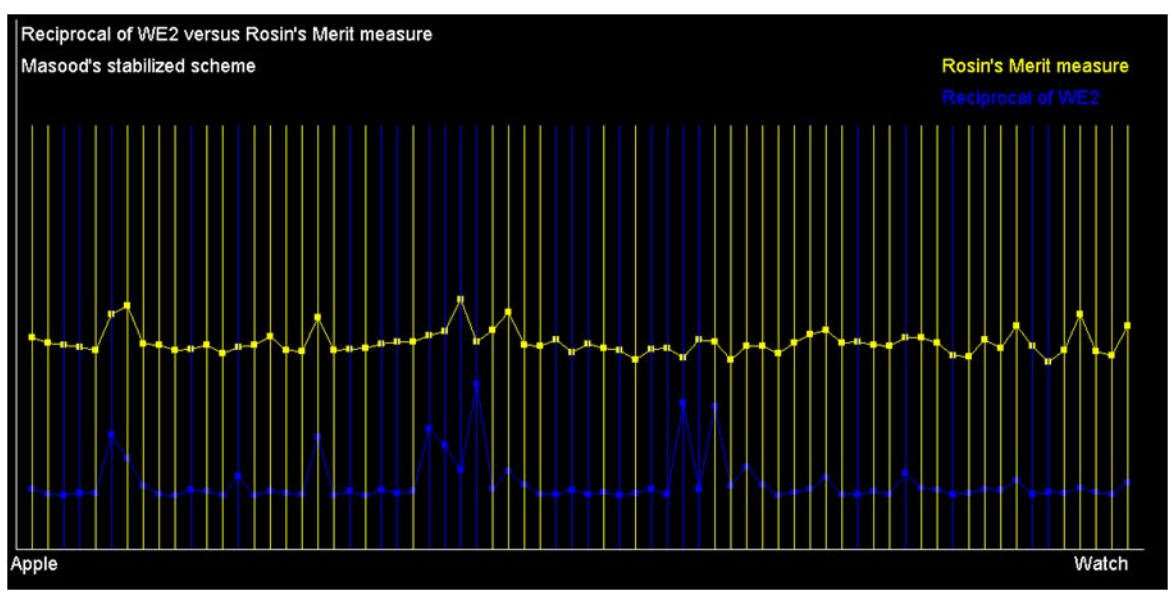

**FIGURE 11** A plot of the reciprocal of $(WE_2)_{suboptimal}$ (in blue) and Rosin's $Merit$ measure (in yellow) using Masood's stabilized scheme. The graph is annotated with vertical lines shown in yellow and blue to facilitate comparison of the two line diagrams. The yellow lines indicate similarity in the pattern of the line diagram, and the blue lines indicate dissimilarity. As there is a mix of yellow and blue lines, the two measures are independent of each other.

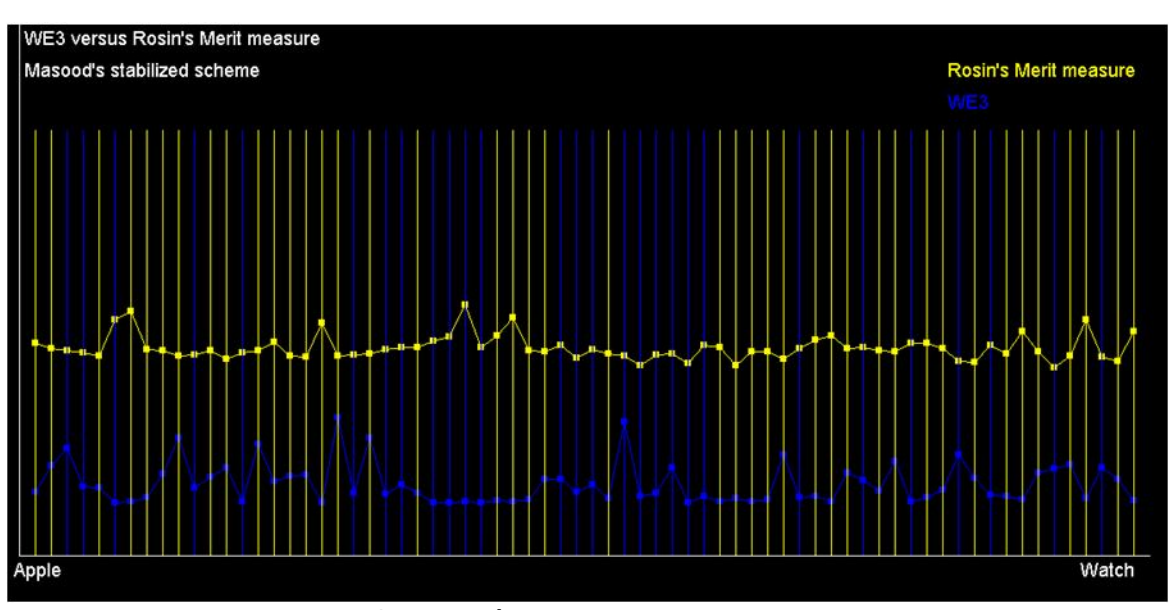

**FIGURE 12** A plot of $(WE_3)_{suboptimal}$ (in blue) and Rosin's $Merit$ measure (in yellow) via Masood's stabilized scheme. The graph is annotated with vertical lines shown in yellow and blue to facilitate comparison of the two line diagrams. The yellow lines indicate similarity in the pattern of the line diagram, and the blue lines indicate dissimilarity. As there is a mix of yellow and blue lines, the two measures are independent of each other.

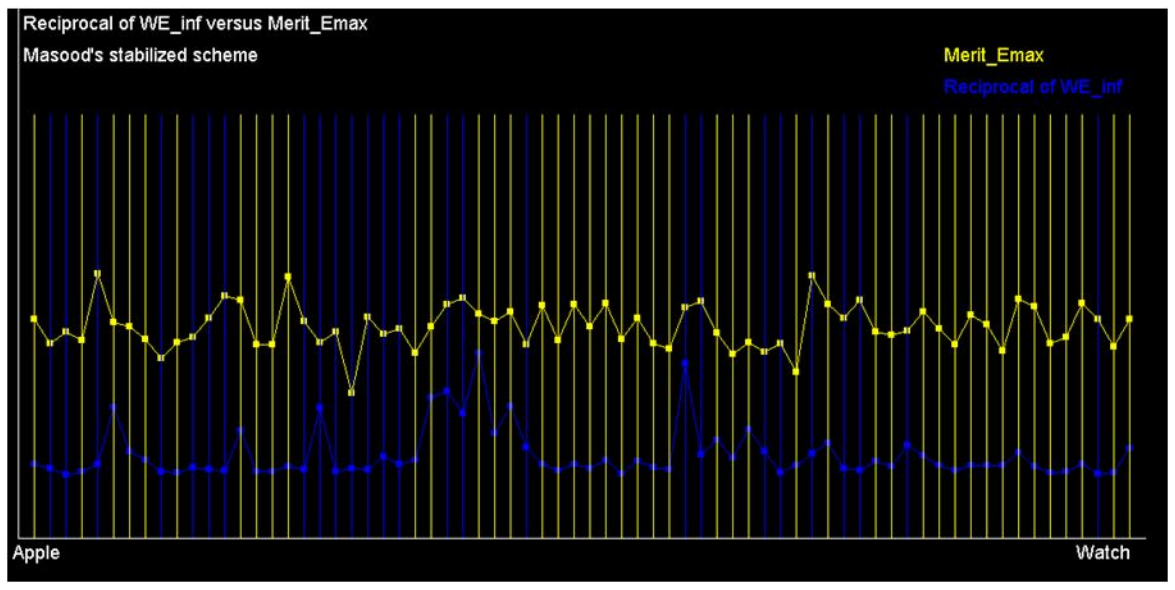

**FIGURE 13** A plot of the reciprocal of $(WE_\infty)_{suboptimal}$ (in blue) and Rosin's $Merit_{E_{max}}$ measure (in yellow) using Masood's stabilized scheme. The graph is annotated with vertical lines shown in yellow and blue to facilitate comparison of the two line diagrams. The yellow lines indicate similarity in the pattern of the line diagram, and the blue lines indicate dissimilarity. As there is a mix of yellow and blue lines, the two measures are independent of each other.

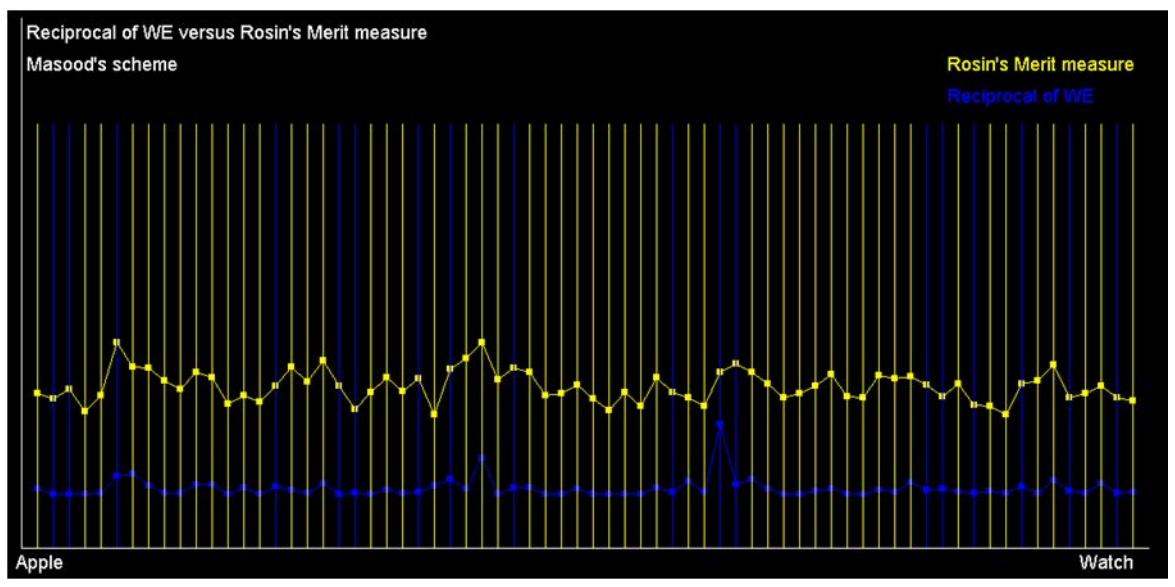

**FIGURE 14** A plot of the reciprocal of $WE_{suboptimal}$ (in blue) and Rosin's $Merit$ measure (in yellow) via Masood's scheme. The graph is annotated with vertical lines shown in yellow and blue to facilitate comparison of the two line diagrams. The yellow lines indicate similarity in the pattern of the line diagram, and the blue lines indicate dissimilarity. As there is a mix of yellow and blue lines, the two measures are independent of each other.

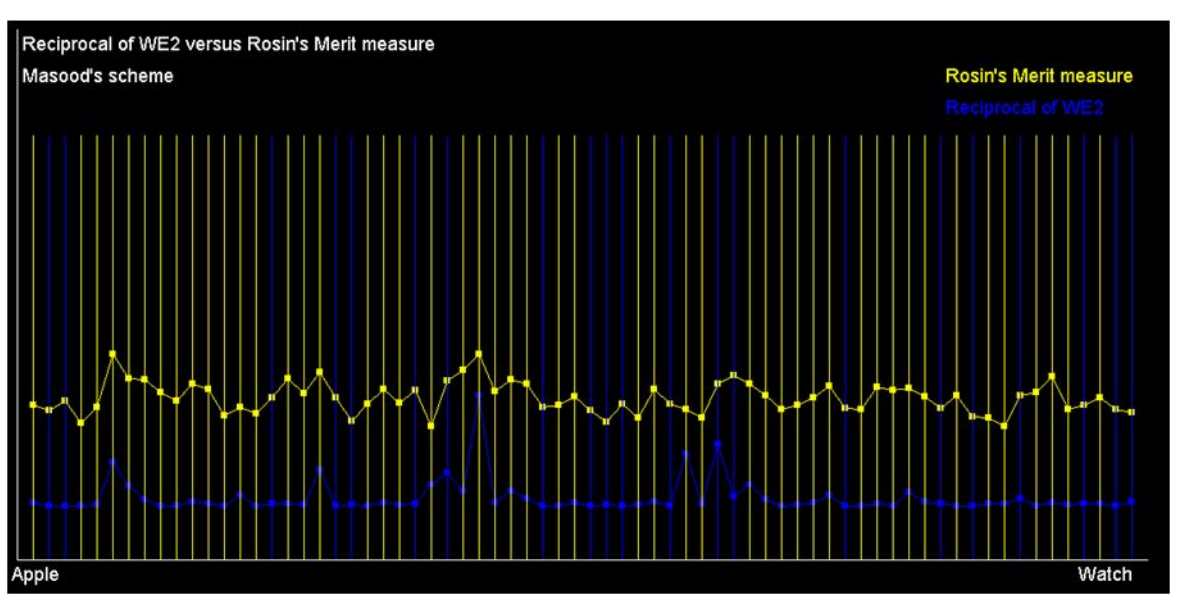

**FIGURE 15** A plot of the reciprocal of $(WE_2)_{suboptimal}$ (in blue) and Rosin's $Merit$ measure (in yellow) via Masood's scheme. The graph is annotated with vertical lines shown in yellow and blue to facilitate comparison of the two line diagrams. The yellow lines indicate similarity in the pattern of the line diagram, and the blue lines indicate dissimilarity. As there is a mix of yellow and blue lines, the two measures are independent of each other.

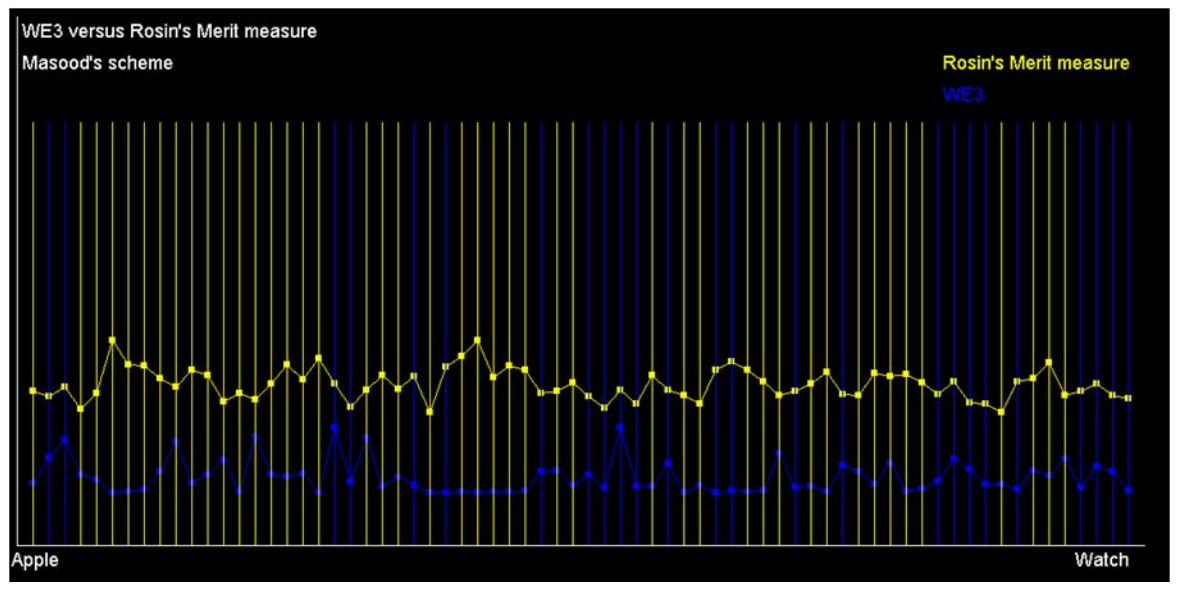

**FIGURE 16** A plot of $(WE_3)_{suboptimal}$ (in blue) and Rosin's $Merit$ measure (in yellow) via Masood's scheme. The graph is annotated with vertical lines shown in yellow and blue to facilitate comparison of the two line diagrams. The yellow lines indicate dissimilarity in terms of rise/fall in the line diagram, and the blue lines indicate similarity. As there is a mix of blue and yellow lines in the graph, the two measures $(WE_3)_{suboptimal}$ (in blue) and Rosin's measure are independent of each other.

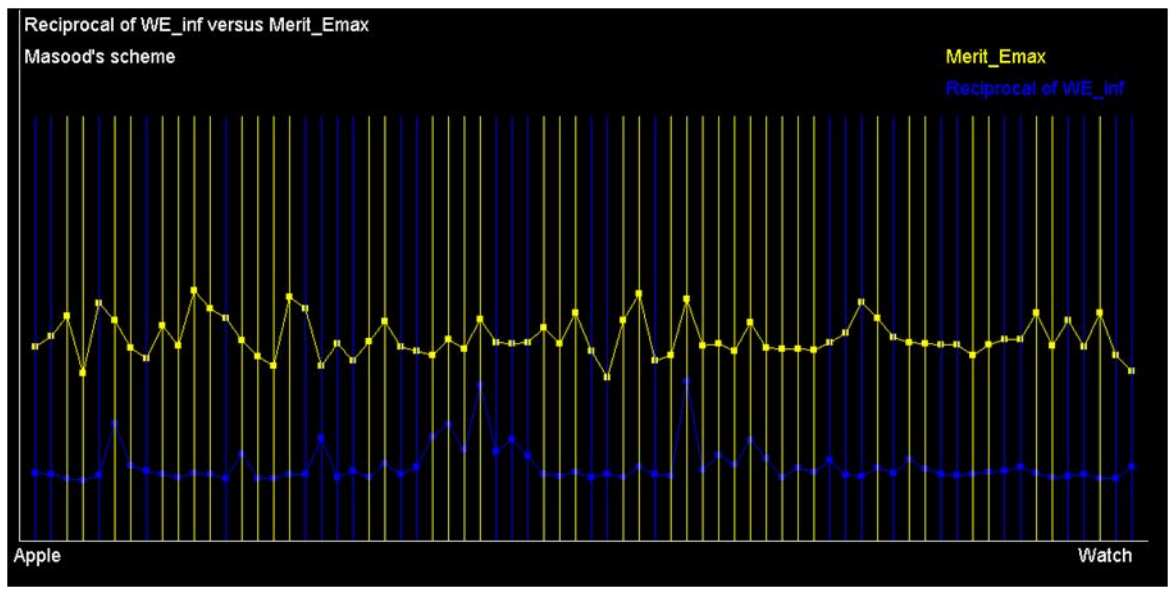

**FIGURE 17 A plot of the reciprocal of the $(WE_\infty)_{suboptimal}$ (in blue) and $Merit_{E_{max}}$ measures (in yellow) via Masood's scheme. The graph is annotated with vertical lines shown in yellow and blue to facilitate comparison of the two line diagrams. The yellow lines indicate similarity in terms of rise/fall in the line diagram, and the blue lines indicate dissimilarity. As there is a mix of yellow and blue lines, the two measures $(WE_\infty)_{suboptimal}$ (in blue) versus $Merit_{E_{max}}$ are independent of each other.**

Fernández-García et al. [29] proposed a measure that is based on the compression ratio ($CR$) and the sum of the square of the errors ($E_2$) and ranked different schemes of polygonal approximation. The measure, called $Fernández - García\ et\ al.\ measure$ in this study, is defined by the arithmetic mean of the reciprocal of the compression ratio and normalized sum of the square of the errors, i.e., $\frac{1}{2}(\frac{1}{CR} + NISE)$, where $NISE = \frac{2}{1+e^{-\frac{\sqrt{E_2}}{D}}} + 1$ and $D = D_1 + D_2$, where $D_1$ is the maximum distance of the curve from its centroid and where $D_2$ is the maximum distance of the curve from the line of minimum inertia. This measure, as it involves $CR$ and $E_2$, is compared with Rosin's $Merit$ measure via the four schemes of polygonal approximation considered in this study. The graphical representations of the comparisons are shown in Figures 18 through 21 in the form of line diagrams, where the blue and yellow line diagrams represent the behavior of $Fernández - García\ et\ al.\ measure$ and Rosin's $Merit$ measure, respectively. The annotated yellow and blue vertical lines present in an interleaving fashion are indicative of the asynchronous behavior of the two line diagrams, establishing that there is no reason to conclude that $Fernández - García\ et\ al.\ measure$ and Rosin's $Merit$ measure are related.

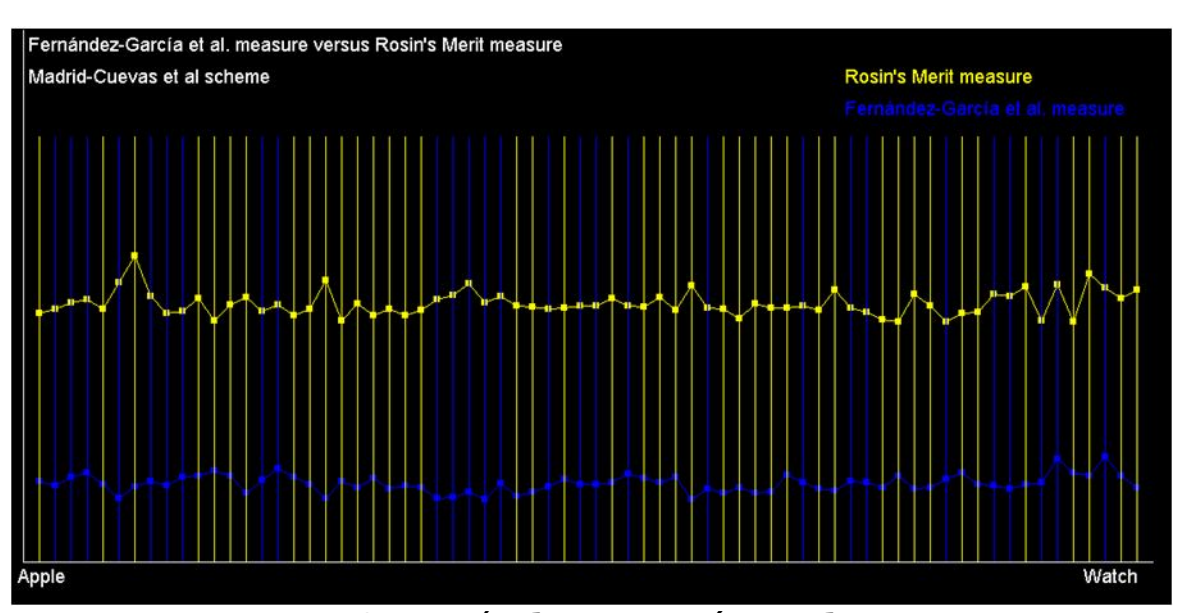

**FIGURE 18 A plot of $Fernández - García\ et\ al.\ measure$ (in blue) versus Rosin's $Merit$ measure (in yellow) via the Madrid--Cuevas et al. scheme. The graph is annotated with vertical lines shown in yellow and blue to facilitate comparison of the two line diagrams. The yellow lines indicate similarity in the pattern of the line diagram, and the blue lines indicate dissimilarity. As there is a mix of yellow and blue lines, the measures are independent of each other.**

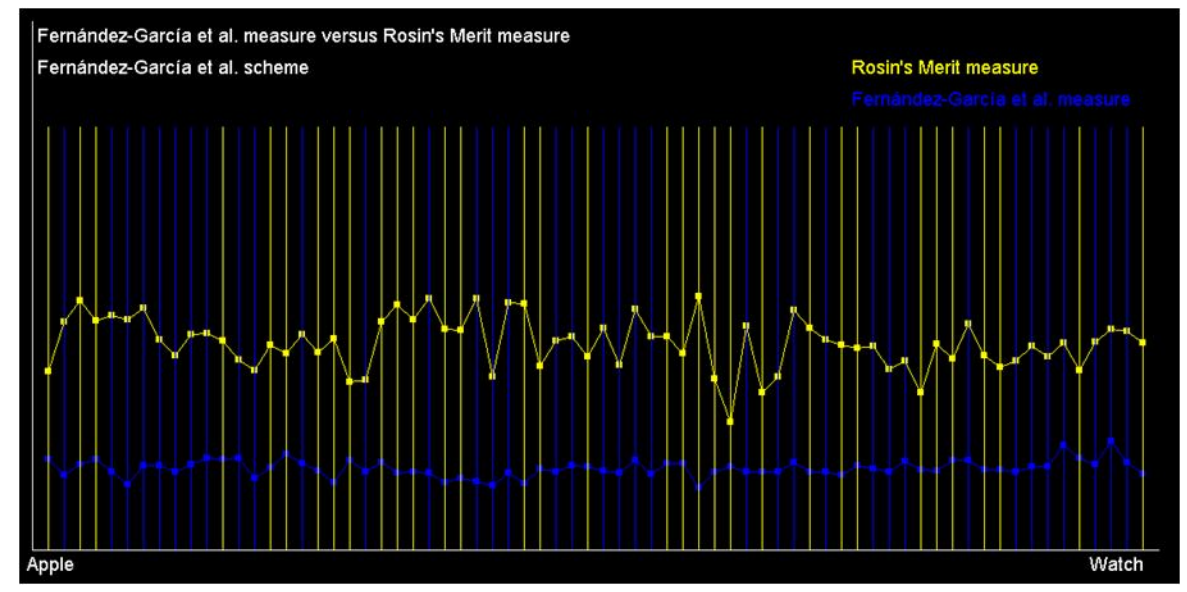

**FIGURE 19 A plot of $Fernández - García\ et\ al.\ measure$ (in blue) versus Rosin's $Merit$ measure (in yellow) via the $Fernández - García\ et\ al.$ scheme. The graph is annotated with vertical lines shown in yellow and blue to facilitate comparison of the two line diagrams. The yellow lines indicate similarity in the pattern of the line diagram, and the blue lines indicate dissimilarity. As there is a mix of yellow and blue lines, the measures are independent of each other.**

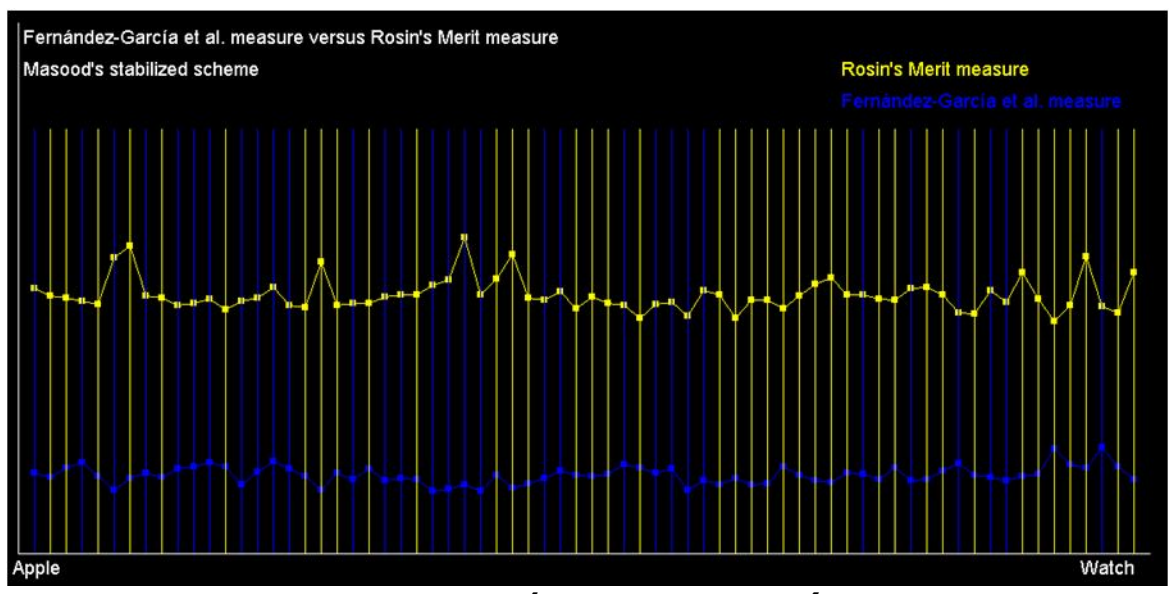

**Figure 20** A plot of $Fernández - García\ et\ al.\ measure$ (in blue) versus Rosin's $Merit$ measure (in yellow) via Masood's stabilized scheme. The graph is annotated with vertical lines shown in yellow and blue to facilitate comparison of the two line diagrams. The yellow lines indicate similarity in the pattern of the line diagram, and the blue lines indicate dissimilarity. As there is a mix of yellow and blue lines, the measures are independent of each other.

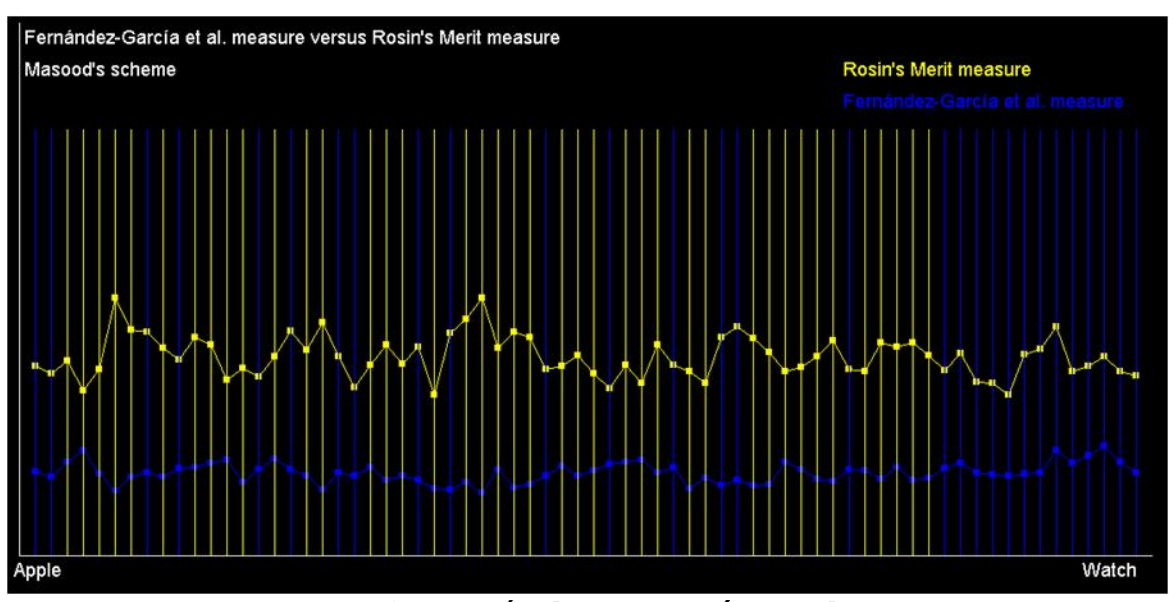

**FIGURE 21** A plot of $Fernández - García\ et\ al.\ measure$ (in blue) versus Rosin's $Merit$ measure (in yellow) via Masood's scheme. The graph is annotated with vertical lines shown in yellow and blue to facilitate comparison of the two line diagrams. The yellow lines indicate similarity in the pattern of the line diagram, and the blue lines indicate dissimilarity. As there is a mix of yellow and blue lines, the measures are independent of each other.

In addition to a graphical analysis of the experimental results, a statistical analysis of the outputs of the experiments is also carried out. Since the objective of this communication is to explore the possibility of a relationship, if any, between the weighted figure of merit (and $Fernández - García\ et\ al.\ measure$) and Rosin's $Merit$ measure along with $Merit_{Emax}$, Pearson's product–moment correlation coefficient between the two types of measures is computed. The data used in the graphical analysis are used for computing the correlation coefficient, and the results are shown in the following table.

| **Scheme of polygonal approximation** | **WE** | **$WE_2$** | **$WE_3$** | **$WE_\infty$** | **Fernández-García et al. measure** |
|---|---|---|---|---|---|
| Madrid-Cuevas et al. | 0.2555 | 0.3164 | -0.4440 | 0.1150 | -0.16267 |
| Fernandez-Garcia et al. | 0.1264 | -0.0248 | -0.1182 | -0.0002 | -0.12038 |
| Masood stabilized | 0.2324 | 0.3127 | 0.1037 | 0.0357 | -0.44544 |
| Masood | 0.5027 | 0.5273 | -0.3042 | 0.1452 | -0.3966 |
| **Average Correlation** | **0.2792** | **0.2829** | **-0.1906** | **0.0739** | **-0.28127** |

**TABLE 1 Correlation coefficient between the weighted figure of merit along with the Fernández-García et al. measure and Rosin's $Merit$ measure and $Merit_{Emax}$**

The row headers of the table show the name of the scheme, and the column headers indicate the different weighted figures of merit and $Fernández - García\ et\ al.$ measures. The cells of the table have the value of the correlation coefficient between Rosin's $Merit$ measure/$Merit_{Emax}$ measure and the weighted figure of merit/$Fernández - García\ et\ al.$ measure. The last row of the table also shows the average value of correlation coefficients over different schemes of polygonal approximation. The table shows that the correlation coefficients are not near unity (negative/positive). Therefore, there is no linear relationship between the two measures.

The nonlinear correlation coefficient is also computed via the distance correlation method proposed in [30]. The results are shown in Table 2 below. The data in the table manifest similar behavior as those in Table 1.

| **Scheme of polygonal approximation** | **WE** | **$WE_2$** | **$WE_3$** | **$WE_\infty$** | **Fernández-García et al. measure** |
|---|---|---|---|---|---|
| Madrid-Cuevas et al. | 0.4729 | 0.4612 | 0.4401 | 0.2138 | 0.3280 |
| Fernandez-Garcia et al. | 0.1264 | 0.2209 | 0.2695 | 0.1697 | 0.3605 |
| Masood stabilized | 0.2379 | 0.2209 | 0.2200 | 0.1697 | 0.3315 |
| Masood | 0.6340 | 0.5492 | 0.4183 | 0.2143 | 0.2918 |
| **Average Nonlinear Correlation** | **0.3678** | **0.3631** | **0.3370** | **0.1919** | **0.3280** |

**TABLE 2 Nonlinear correlation coefficient between the weighted figure of merit along with the Fernández-García et al. measure and Rosin's $Merit$ measure and $Merit_{Emax}$**

There are various statistical measures that facilitate the study of relationships between sets of data. The Pearson correlation coefficient and distance correlation are used here because the data are quantitative in nature. The other statistical measures used to compute the degree of association between sets of data are Spearman's rank correlation coefficient and the Kendall's $\tau$ $(Tou)$ and $\chi^2$ $(Chi - square)$ tests; however, these tests are not appropriate for this study. Spearman's rank correlation coefficient and Kendall's $\tau$ are used for ordinal data, whereas $\chi^2$ is used to determine whether the observed value and the estimated value of an attribute (correlation addresses two attributes) are associated with each other at a specific level of significance. Notably, the line diagrams presented in the previous discussion support the fact that the measures are independent, which is further strengthened by the correlation coefficient.

The results of the theoretical analysis, experiments and statistical analysis indicate that Rosin's measure and weighted figure of merit are independent of each other. It is not possible to infer one from the other. If a suboptimal scheme is found to be better than others when a weighted figure of merit is used as a metric, then the same conclusion cannot be drawn when Rosin's measure is used. Since Rosin's measure is time-consuming to compute, researchers are tempted to use a weighted figure of merit. However, there are multiple reasons for using Rosin's measure instead of the weighted figure of merit. Rosin's measure is derived analytically and uses an optimal scheme as a base to assess a suboptimal scheme, whereas the weighted figure of merit is ad hoc in nature and does not take into account optimal approximation to assess a suboptimal scheme. When one is looking for an alternative to a measure, the latter should behave in a similar manner as the former. Although Rosin's measure is known to produce a high value (indicating a good approximation) for a polygonal approximation containing break points only and by approximation consisting of three vertices only, these approximations are trivial approximations. Any metric used to assess nontrivial polygonal approximations should have a sound mathematical basis and should behave synchronously with Rosin's measure. In the absence of any such measure, when suboptimal schemes for polygonal approximation are compared, one needs to use Rosin's measure. Although it is time consuming to compute Rosin's measure because of its involvement with the optimal scheme, the time is consumed during testing of a polygonal approximation scheme but not in its usage in subsequent computer vision applications.

## VI CONCLUSION

The goodness of fit of a suboptimal scheme for polygonal approximation is usually measured through its comparison with an optimal scheme. The optimal schemes for polygonal approximation are computationally expensive, leading to a high testing time to measure the goodness of fit of a suboptimal scheme. This is why researchers have used a weighted figure of merit instead of Rosin's measure to compare various suboptimal schemes. However, it is found in this communication through theoretical analysis, experiments and statistical analysis that a weighted figure of merit and its alternative, such as the Fernández-García et al. measure, cannot be a substitute for Rosin's measure because the two measures are independent of each other. Any measure of goodness for the polygonal approximation introduced in the future should be assessed in line with Rosin's measure. The objective of this communication is not to compare a weighted figure of merit and Rosin's measure to determine which one is better than the other as a measure of the goodness of fit of a polygonal approximation scheme; rather, it is observed here through this investigation that one cannot use a weighted figure of merit instead of Rosin's measure to sidestep its computational load. As future research in this direction, it may be desirable to discover a measure that is computationally more efficient than Rosin's measure, is in sync with it in measuring the goodness of fit of a scheme and has a sound mathematical basis.